\documentclass[letterpaper, 10 pt, conference]{ieeeconf}  

\IEEEoverridecommandlockouts                              

\usepackage{color}
\usepackage{graphics} 
\usepackage{epsfig} 
\usepackage{times} 
\usepackage{amsmath} 
\usepackage{amssymb}  
\usepackage{multirow}

\title{\LARGE \bf
A Rapid Pattern-Recognition Method for Driving Styles Using Clustering-Based Support Vector Machines
}

\author{Wenshuo Wang$^{1}$ and Junqiang Xi$^{2}$
\thanks{$^{1}$Wenshuo Wang is a Ph.D. candidate in Mechanical Engineering, Beijing Institute of Technology, Beijing, China, and studying in the Vehicle Dynamics \& Control Lab, UC Berkeley.
        { (wwsbit@gmail.com)}}%
\thanks{$^{2}$Junqiang Xi is a Professor in Mechanical Engineering, Beijing Institute of Technology, Beijing, China.
        { (xijunqiang@bit.edu.cn)}}%
}

\begin{document}

\maketitle


\begin{abstract}

A rapid pattern-recognition approach to characterize driver's curve-negotiating behavior is proposed. To shorten the recognition time and improve the recognition of driving styles, a $ k $-means clustering-based support vector machine ($ k $MC-SVM) method is developed and used for classifying drivers into two types: aggressive and moderate. First, vehicle speed and throttle opening are treated as the feature parameters to reflect the driving styles. Second, to discriminate driver curve-negotiating behaviors and reduce the number of support vectors, the $ k $-means clustering method is used to extract and gather the two types of driving data and shorten the recognition time. Then, based on the clustering results, a support vector machine approach is utilized to generate the hyperplane for judging and predicting to which types the human driver are subject. Lastly, to verify the validity of the $ k $MC-SVM method, a cross-validation experiment is designed and conducted. The research results show that the $ k $MC-SVM is an effective method to classify driving styles with a short time, compared with SVM method.

\end{abstract}

\section{Introduction}

To design an intelligent and human-centered control system \cite{wang15} that adaptively adjusts relevant parameters in time to meet the human driver's needs and to provide a basic control law for the advanced vehicle dynamics control system \cite{bich10}\cite{okam10} or driver assistance system \cite{wang13}\cite{angk11}, driver behaviors,  driving styles or characteristics should be recognized and predicted.  For example, to  improve vehicle's fuel economy and reduce the emission, we can design different control strategies for driving styles. To achieve these goals, recognition and prediction of driving styles and characteristics precisely is the primary work. Drivers and their factors have been discussed from the viewpoint of application in vehicle dynamics \cite{ploc07}\cite{wws15}, physical attributes of human drivers, and modeling driver \cite{maca03}\cite{wang14}. For the recognition and prediction of driving characteristics or driver types, including physical characteristics/states (e.g., fatigue, drunk, and drowsiness), psychical characteristics (e.g., nervous, relaxed) and driving styles (e.g., aggressive, moderate), a lot of investigations have been conducted in recent years.

In general, the basic idea to identify and predict driving behaviors or styles is based on driver model, called {\it indirect} or {\it model-based method}. The model-based method, firstly, requires to establish a driver model that can describe driver's basic driving behaviors, i.e., lane keeping, lane change, and obstacle avoidance, etc. Then identify or extract driving characteristics based on the proposed driver model. Many stochastic process theories were applied to the recognition of driving behaviors. Hidden Markov Model (HMM), as a simple dynamic Bayesian network, can identify the underlying relationship between observations and states, and then be widely utilized to model and predict driver's states \cite{tade14}, driving behaviors \cite{gade14}\cite{mitr05}. In \cite{gade14}, the driver-vehicle system was treated as a hybrid-state system, and  the HMM was used to estimate driver's decisions when driving near intersections.  In \cite{sund13}, a probabilistic ARX model was utilized to predict the human driver's behavior and classify the driving styles, i.e., the aggressive and normal driving. To mimic and model the uncertainty of driver's behavior, a stochastic switched-ARX (SS-ARX) model was developed and adopted by Akita et al. \cite{akita07} and Sekizawa et al. \cite{seki07}. 

The other recognition method to extract and identify driver's driving characteristics, called {\it direct method} . The basic process of the direct method is to directly analyze the driving data (e.g., vehicle speed, steering wheel angle, throttle opening, etc.) using pattern-recognition or data-analysis method without establishing relevant driver models. For recognition of driving skills, Zhang et al. \cite{zhang10} proposed a direct pattern-recognition approach based on three recognition methods, i.e., multilayer perception artificial neural networks (MLP-ANNs), decision tree, and support vector machines (SVMs). The coefficients of discrete Fourier transform (DFT) of steering wheel angles were treated as the discriminant features. In \cite{higg15}, relationships between driver state and driver's actions were investigated using the cluster method with eight state-action variables. For different driving patterns of drivers, the state-action clusters were different, thus segmenting driver into different patterns. 

In this paper, a direct method to recognize {\it driving styles} is proposed by combining $ k $-means clustering ($ k $-MC) method and SVM method together. In terms of traditional recognition or classification method such as SVM, ANNs, and ARX, most of them take a long time to calculate the reasonable results, especially for not linearly separable issues. 
In order to develop an efficient, time-saving, and direct method to recognize driving styles, the $ k $MC-SVM method is adopted. There are three main steps of this research effort: 1) {\it Clustering}. To extract the discrimination features that can distinctly reduce the number of support vectors, the driving data is clustered into $ K $ subsets using the $ k $-MC method. 2) {\it Training and generating a hyperplane}. The hyperplane that can disparate driving styles into two types is generated using the $ k $MC-SVM method, to predict the new input data to which category a driver is subjected. 3) {\it Experimental verification}. A cross-validation experiment is designed to recognize driving styles and to present the benefits of the proposed method. 

Following the overview in the first section of this paper, Section II presents the basic framework of the $ k $MC-SVM method. To show the advantages of $ k $MC-SVM method, the experiments and driving simulator to collect driving data is shown in section III. Subsequently, the evaluation of recognition performance is discussed in section IV.

\section{Pattern Recognition Method}
\begin{figure}[thpb]
\centering
\includegraphics[scale=0.25]{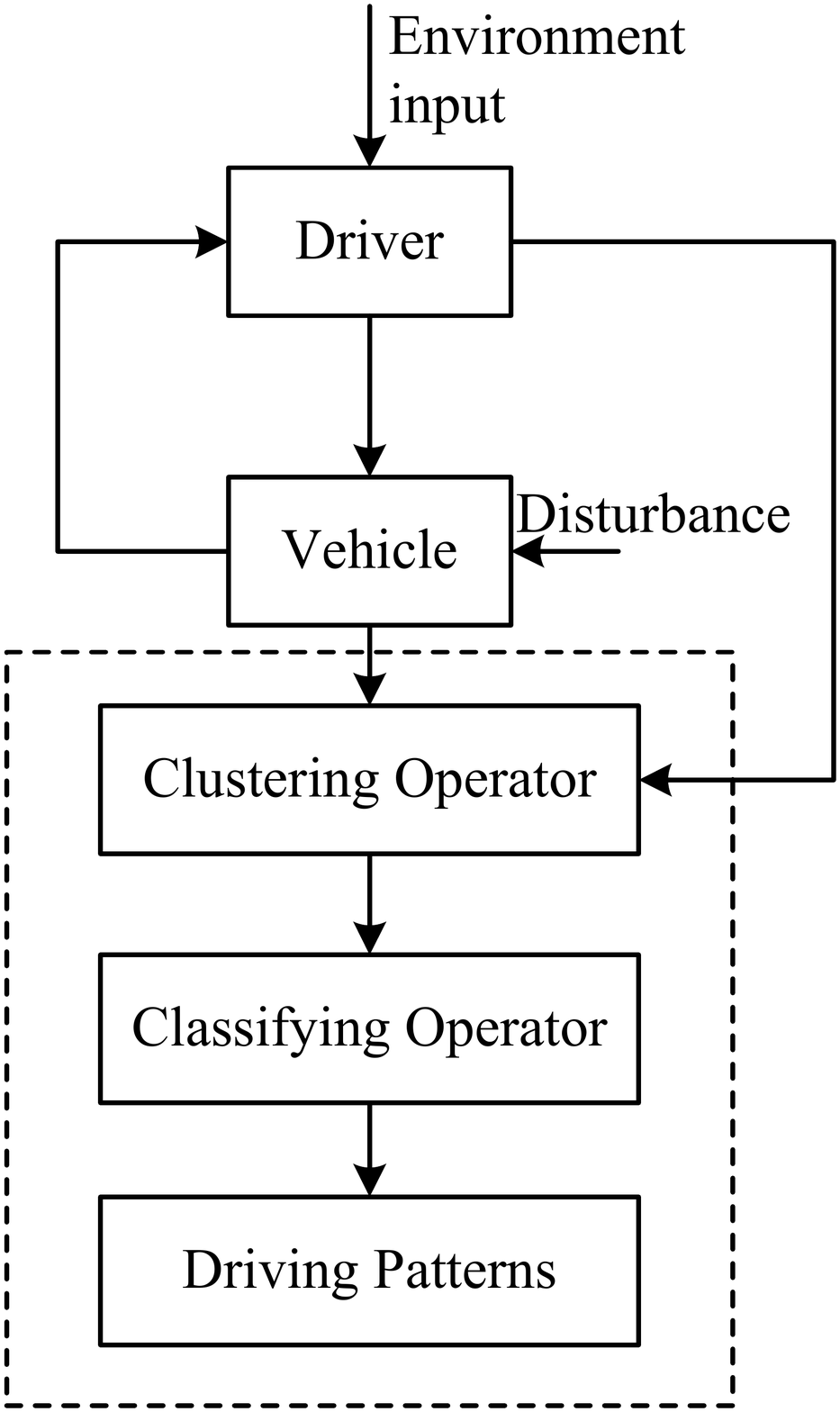}
\caption{The schematic diagram of the proposed pattern-recognition method for driving styles.}
\label{figure1}
\end{figure}

In this section, the parameters selected for $ k $MC-SVM method, the SVM and $ k $-MC are discussed. The structure of proposed recognition method is shown as Fig. \ref{figure1}. 

\subsection{Parameters Selection}
To characterize driving styles when drivers pass a curve road, the longitudinal speed $ v_{x} $ and throttle opening $ \alpha $ are treated as the feature vector parameter $ \mathbf{x}=(v_{x},\alpha) $. 

\subsubsection{Longitudinal speed} The longitudinal vehicle speed can reflect the driver's driving preferences. For instance, if a driver prefers high vehicle speed when passing a curve road, the driver is treated as an aggressive driver. Inversely, we will treat the driver as a moderate type. Therefore, the vehicle speed driver selects is treated as one of the feature parameters.

\subsubsection{Throttle opening} The longitudinal acceleration of the vehicle can also represent the driving styles, i.e., the higher acceleration of vehicle, the more aggressive the driver would be treated as. The throttle opening is directly controlled by the human driver and highly related to the acceleration. Therefore, the throttle opening is also selected as one of the feature parameters.

To describe the driving styles precisely, the combination of longitudinal vehicle speed and throttle opening is conducted. Namely, a low (high) speed with a large (small) throttle opening denotes more aggressive, and a high (low) speed with a large (small) throttle opening denotes less aggressive. 
Therefore, the mapping between driving styles and driving data can be written as $ f: \mathcal{X} \rightarrow \mathcal{Y} $, where $ \mathcal{X}:=\{ \mathbf{x}_{i} $, $ i=1,2,\dots,n \} $ is a set of all collected data $ \mathbf{x}_{i} $, $ \mathcal{Y}:= \{ \boldsymbol{y}_{i}, i=1,2,\dots,n \} $. Our goal is to train and learn the mapping $ f: \mathcal{X} \rightarrow \mathcal{Y}  $ that can accurately and rapidly classify new data sets.

\subsection{Support Vector Machines}
SVM, as a tool of solving problems in classification, regression, and novelty detection, becomes popular in recent years and has been widely involved in voice or speaker detection \cite{hatch06}, image processing \cite{cusa03}, human action detection \cite{schu04}, etc. The SVM method is discussed as follows.

\subsubsection{Problem Formulation Based on SVM}

An important property of SVM is that determining the model parameters is equal to solve a convex problem, guaranteeing the global optimum  \cite{bishop06}. 
In this work, the classes of driving patterns are not linearly separable, i.e., it belongs to the situation of overlapping class distribution.
Considering the case in which there are two driving-patterns in this paper that are not linearly separable in $ q $ dimension space. For each of the $ n $ training cases, there is a vector, denoting $ \boldsymbol{\eta}_{i} $, represents the feature parameters with corresponding target value, denoting $ y_{i} $. Therefore, the training data could be written as $ \{\boldsymbol{\eta}_{i},y_{i} \} $, $ i=1,2,\dots,n $, $ y_{i} \in \{ -1,1 \}$. Here, the aim is to generate a hyperplane in a high-dimension space to separate the two classes accurately. To separate classes as accurately as possible for the inseparable situation, the slack variable, $ \xi_{i}\geqslant 0 $, where $ i=1,2,\dots,n $, is introduced. For the new input data $ \tilde{\boldsymbol{\eta}}  $, its target value can be calculated by 

\begin{equation}
g(\boldsymbol{\eta})=\mathbf{w}^{\top}\phi(\boldsymbol{\eta})+b+\xi
\end{equation}
where $ \phi(\boldsymbol{\eta}) $ is a feature-space transform, $ b $ is a bias parameter. The slack variable, $ \xi $, allows some of the training data to be misclassified, generating a {\it soft margin}. Then, the objective function can be described as follows:

\begin{equation}
\min \quad J=C\sum_{i=1}^{n}\xi_{i}+\frac{1}{2}\parallel \mathbf{w} \parallel ^{2}
\end{equation}
where the parameter $ C $ is used to adjust the trade-off between slack variable penalty and the margin. Transform Equation (2) into the Lagrangian form:

\begin{equation}
L(\mathbf{a})=\sum_{i=1}^{n}a_{i}-\frac{1}{2}\sum_{i=1}^{n}\sum_{j=1}^{n}a_{i}a_{j}y_{i}y_{j} \mathcal{K} (\boldsymbol{\eta}_{i},\boldsymbol{\eta}_{j})
\end{equation}
with constraints as follows: $\sum_{i=1}^{n}a_{i}y_{i}=0,\ 0\leqslant a_{i} \leqslant C$, where $ \mathbf{a}=\{a_{i}\}, i=1,2,\dots,n $ is the Lagrangian multiplier, $ \mathcal{K}(\cdot,\cdot) $ is the kernel function. Here, the kernel function is positive definite and Gaussian kernel function is selected and presented as follows:

\begin{equation}
\begin{split}
\mathcal{K}(\boldsymbol{\eta}_{i},\boldsymbol{\eta}_{j})= & \phi(\boldsymbol{\eta}_{i})^{\top} \phi(\boldsymbol{\eta}_{j}) \\
= & \exp \left\lbrace \frac{-\parallel \boldsymbol{\eta}_{i} - \boldsymbol{\eta}_{j} \parallel ^{2}}{2\sigma^{2}} \right\rbrace \\
= & \exp \left\lbrace -\gamma \parallel \boldsymbol{\eta}_{i} - \boldsymbol{\eta}_{j} \parallel ^{2} \right\rbrace 
\end{split}
\end{equation}
where $ \gamma=1/(2\sigma^2) $ is the Gaussian kernel parameter.

\subsubsection{Parameters Determination} The determination of SVM parameters ($ C,\gamma $) is discussed in this section. Our goal is to find the optimal parameters ($ C_{opt},\gamma_{opt} $), allowing the classifier to accurately predict the unknown data. For the highly linearly-inseparable set of observations, it may not be the best for parameters that could  highly separate the training data into an accuracy region, which may lead to the problem of over-fitting the training data. To overcome this issue, the cross-validation procedure is adopted.

Here, the {\it cross-validation} and {\it grid-search} method are adopted to determine the optimal parameters ($ C_{opt},\gamma_{opt} $). According to \cite{hsu03}, the exponentially growing sequences of $ C $ and $ \gamma $ shows a better performance of identifying the optimal parameters. Therefore, parameters $ C $ and $ \gamma $ are all selected by the following form:

\begin{equation}
\begin{split}
\boldsymbol{C}      =& \left\lbrace C \mid C=c^{M} \right\rbrace \\
\boldsymbol{\gamma} =& \left\lbrace \gamma \mid \gamma=r^{-(2N+1)} \right\rbrace 
\end{split}
\end{equation}
where $ c,r,M $ and $ N \in \mathbb{R} $ and, here, $ c=r=2 $ and $ \{M\},\{N\} $ are the arithmetic sequences with initial value $ -5 $, end value $ 10 $ and interval value $ 1 $. For SVM training, the initial value ($ C_{init}, \gamma_{init}$) is set as ($ 2^{0},2^{-1} $). After training the data set, we get the optimal value ($ C_{opt},\gamma_{opt} $) is ($ 2^{7},2^{-9} $), i.e., $ M=7 $ and $ N=4 $.
Based on above descriptions, the driving patterns are classified using $ k $MC-SVM method and the optimal separating hyperplane is produced, as shown in Fig. \ref{figure5} and Fig. \ref{figure5-1}. 

\subsection{$ k $-means Clustering}
As one of the direct methods to analyses and extracts feature parameters from the large volume of raw data, the aim of clustering method is to provide {\it objective} and {\it stable} classifications \cite{ever11}. Objective in the case means that the analysis of the same set of drivers using the same sequence of numerical methods results in the same classification. Stable denotes that the classification remains the same under a wide variety of additions of drivers.

Assume that a set $ Q $ of $ n $ training examples ($\mathbf{x}_{1},y_{1}$) $,\dots, (\mathbf{x}_{n},y_{n}) \in \mathcal{X} \times \mathcal{Y}$ is recorded. $ \mathcal{X} $ is the set of all driving data sets and $ \mathcal{Y} $ is the set of all labels that drivers subject to. Generally speaking, the driving data would be thousands of data points ($ \mathbf{x}_{i}, y_{i}$) and they are highly overlapped data sets. To reduce the number of support vectors and separate the raw feature parameters for different types driving patterns, the $ k $-MC method is applied. For any data point ($ \hat{\mathbf{x}}_{i}, y_{i}$) $ \in \mathcal{X} \times \mathcal{Y} $,  the remaining data sets would be clustered and generating a new cluster, if the remaining data could meet the following condition

\begin{equation}
\begin{split}
& \parallel (\hat{\mathbf{x}}_{i},y_{i})-(\mathbf{x}_{j},y_{j}) \parallel^2 \leqslant r  \\
& i\neq j \, \mbox{and} \, i,j=1,2,\dots,n \\
& i,j \leqslant n, \, r \in \mathbf{R}^+
\end{split}
\end{equation}
where $ r $ is a constant parameter that determine the center-radius of new clusters, $ y_{i} $ and $ y_{j} $ must have the same label. Therefore, the application of $ k $-MC could refine the raw data into a new set. $ k $-MC is used to partition the raw data sets $ \{\mathbf{x}_{i},y_{i} \} $ into $ K $ ($ K \leqslant n $) clusters, forming a set $ \mathcal{O}:=\{(\hat{\mathbf{x}}_{l},y_{l}),\,l=1,2,\dots,K\} $. The $ (\hat{\mathbf{x}}_{l},y_{l}) $ is the subset of set $ \mathcal{O} $. In this paper, the $ k $-MC is calculated by optimizing the following objective function:

\begin{equation}
\arg\,\min_{\mathcal{O}} \sum_{j}^{K}\sum_{\mathbf{x} \in \hat{\mathbf{x}}_{j} }\parallel (\mathbf{x},y_{j})-(\boldsymbol{\eta}_{j},y_{j}) \parallel ^{2}
\end{equation}
where $ (\boldsymbol{\eta}_{j},y_{j}) $ is the mean of point in set $ (\hat{\mathbf{x}}_{j},y_{j}) $, $ K\leqslant n/2 $, and $ n $ is the number of observations. The adopted clustering algorithm can be found in \cite{mack03}. 

\subsection{Training Results Analysis}
\begin{figure}[thpb]
\centering
\includegraphics[scale=0.6]{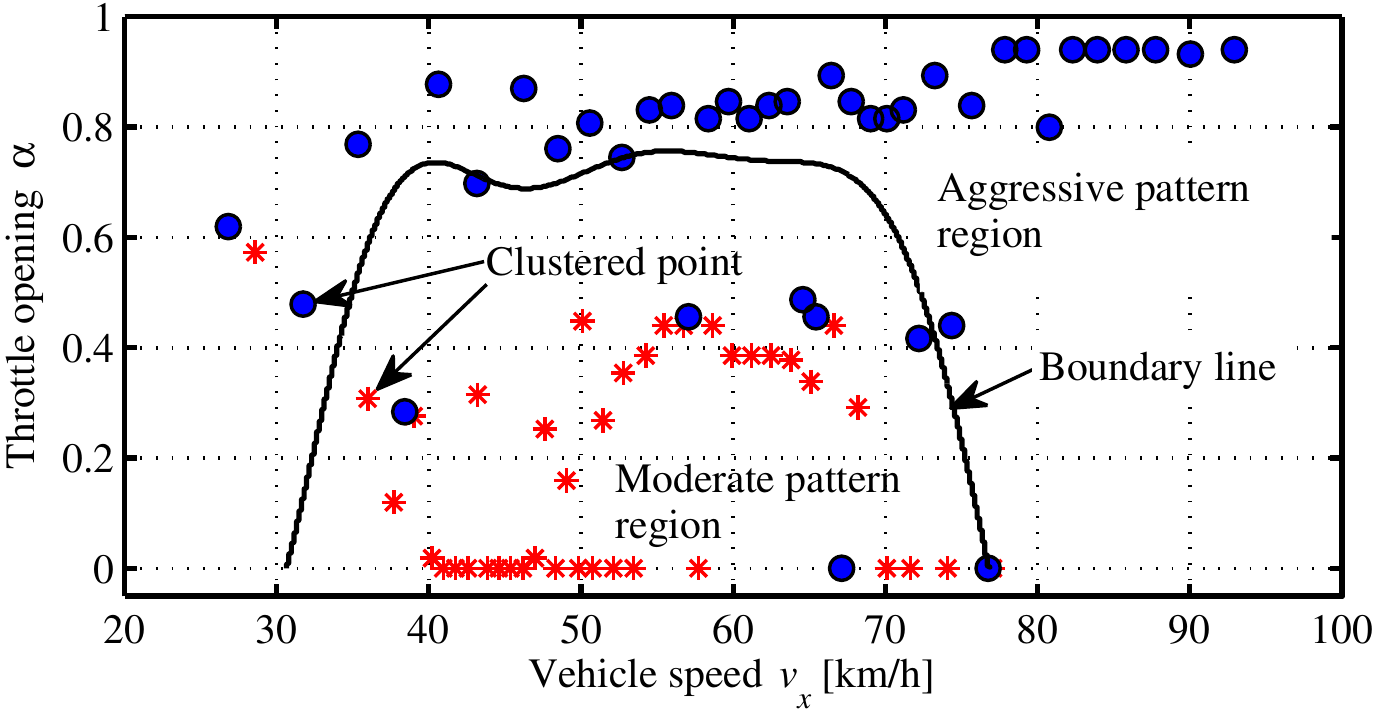}
\caption{The classifying results using $ k $MC-SVM method for driver $ A_1 $ and $ B_1 $. Blue dot: the aggressive driver; red star: the moderate driver.}
\label{figure5}
\end{figure}

Two typical drivers are discussed in the training results.
For moderate driver $ A_{1}  $ in Fig. \ref{figure5}, he prefers lower speed ($\lesssim 40 $ km/h) with smaller throttle openings ($\lesssim0.6 $) when passing a curve road. When vehicle speed $v_{x}\in[40, 65] $ km/h, the moderate driver prefers a  smaller throttle opening ($\lesssim 0.6 $).  When the vehicle speed reaches to certain threshold ($\approx 65 $ km/h), the moderate drive prefers less opportunities in the speed range, $ [65, 80] $ km/h,  than the aggressive driver. According to the Fig. \ref{figure5}, the moderate driver rarely drives the vehicle with speed of larger than $ 80 $ km/h. For aggressive driver $ B_1 $ in Fig. \ref{figure5}, they prefer a larger throttle opening most of the case, though there are some times that the small throttle opening is selected. For the vehicle speed, the aggressive driver prefers has less opportunity to drive vehicle at speed of the range of $ [30, 50] $ km/h and have a higher opportunity to drive at speed range of $ [75,95] $ km/h, in which the moderate driver scarcely does.

\begin{figure}[thpb]
\centering
\includegraphics[scale=0.65]{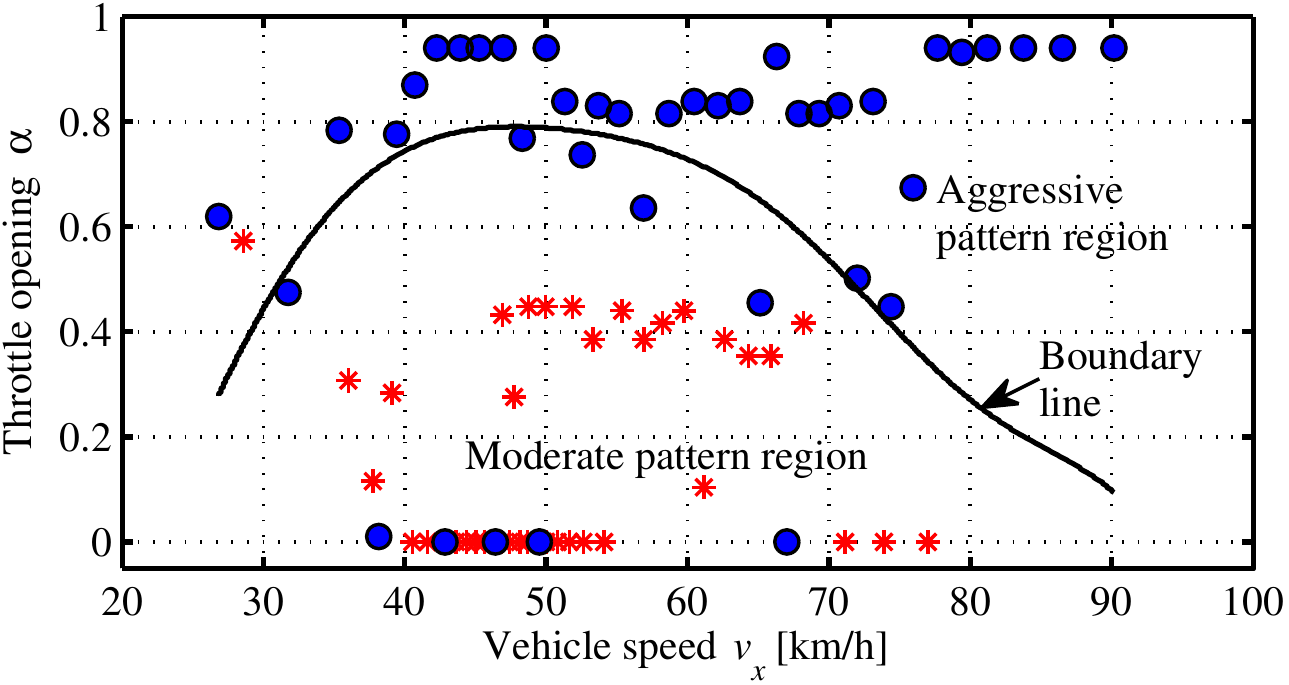}
\caption{The classifying results using $ k $MC-SVM method for driver $ A_2 $ and $ B_2 $. Blue dot: the aggressive driver; red star: the moderate driver.}
\label{figure5-1}
\end{figure}

For moderate driver $ A_2 $ in Fig. \ref{figure5-1}, he prefers to drive vehicle with speed range of $ [35, 75] $ and a small throttle opening ranging from $ 0 $ to $ 0.5 $. For the aggressive driver $ B_2 $, he prefers a lager throttle opening than the moderate driver though the vehicle has a lower speed in $ [25,50] $ km/h, which means that the aggressive driver prefers a lager acceleration.  When $ v_{x} \gtrsim 50 $ km/h, the aggressive driver prefers to chose a lager throttle opening than the moderate driver. Specially, according to Fig. \ref{figure5-1}, we can conclude that the aggressive driver is incline to a higher speed ($ >80 $ km/h) than the moderate driver when they drive on the curve road.

\section{Experiments in Driving Simulator}
In this section, the driver simulator and the collection of the training data are discussed. 

\subsection{Driving Simulator} 
\begin{figure}[thpb]
\centering
\includegraphics[scale=0.55]{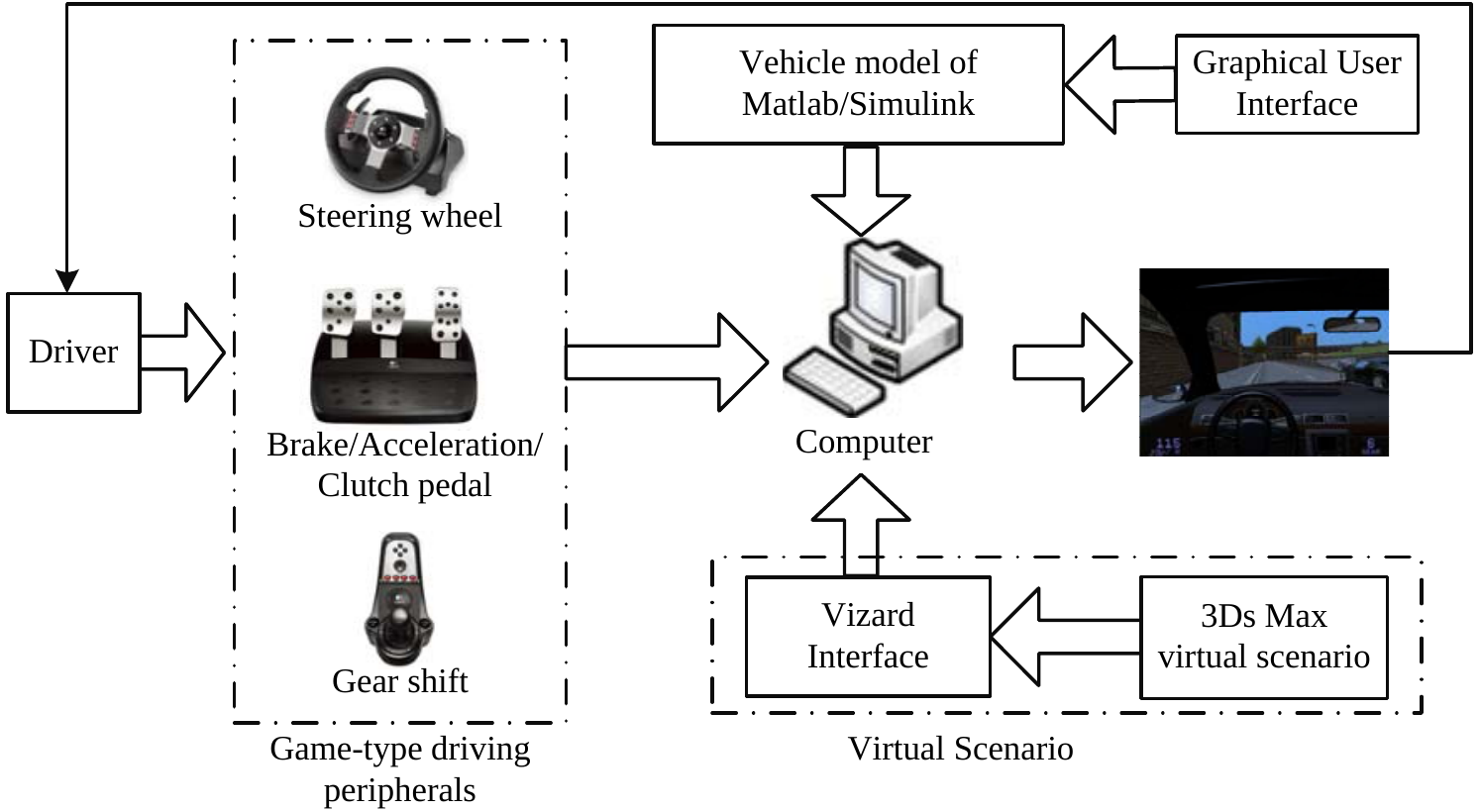}
\caption{The schematic diagram of driving simulator.}
\label{figure6}
\end{figure}

\begin{figure}[thpb]
\centering
\includegraphics[scale=0.6]{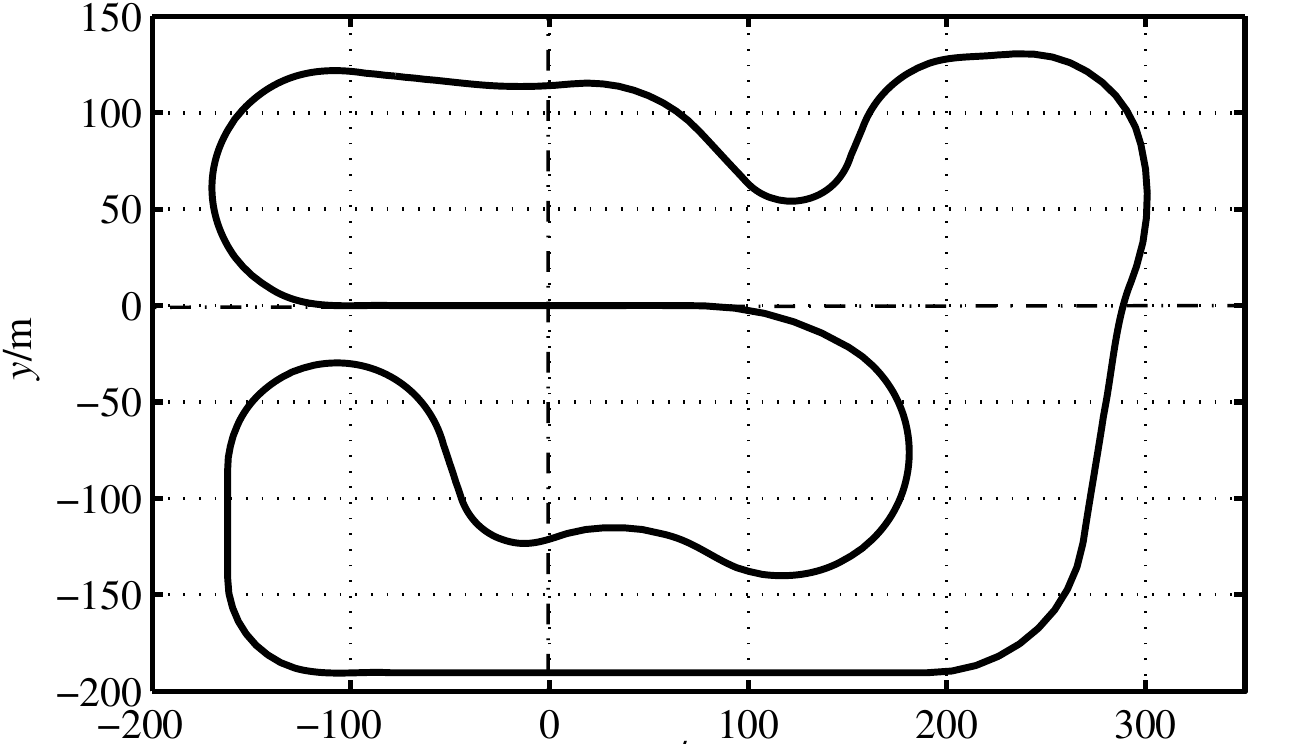}
\caption{The outline of road model for experimental data collection.}
\label{figure7}
\end{figure}

All the experimental data are collected in a driving simulator. The driving simulator consists of five main parts (Fig. \ref{figure6}). The game-type driving peripherals are used to collect the driver's operating signals, i.e., steering wheel angle, brake pedal displacement, throttle opening. The virtual scenarios, including the vehicle, roads, and driving facilities, are designed through 3Ds Max software and Vizard software. The vehicle dynamics model is built using Matlab/Simulink, and a 2-DOF vehicle model is used. The driving environment is considered normal and the road friction $ \mu $ is set as $ 0.9 $. Fig. \ref{figure7} shows the road files we designed. 

\subsection{Training-Data Collection and Experiments}
The driving simulator could record a list of driving parameters such as vehicle speed, vehicle position, steering wheel angle, braking pedal displacement, throttle opening, at a sample frequency of 50 Hz. In the experiment, every driver should be familiarized with the test course and the driving simulator before running an experiment.
Two typical types of the human driver, including four aggressive and four moderate, are involved. To rule out other factors such as the age and gender, all participants are male and aging between 22 $ \sim $ 27 years. Before running the tests, the driving pattern for each driver is determined by a questionnaire way. For each participant, the participant drives vehicle in the simulator more than ten times and every driver is manually labeled the driving pattern such as aggressive or moderate driver before running.

\section{Recognition Performance Evaluation}

The assessment method for the proposed recognizer, i.e., $ k $MC-SVM, is described in this section and the testing results are presented and discussed.

\subsection{Evaluation Method}

To evaluate recognition performance of the proposed recognition method, a well-known evaluation scheme, called {\it cross-validation}, is utilized. For all recorded data from the driving simulator, they are divided into $ z $ subsets of equal size. Then, $ p $ ($ p<z $, $ p \in \mathbb{N}^{+} $) arbitrary subsets of all data sets are used to assessing the performance of classifier trained by the rest $ z-1 $ subsets. This method is called Leave-$ p $-out Cross-validation (L$ p $O-CV). Here, the value of $ p $ is set as $ 1 $. The accuracy of driving-pattern recognizer is defined as:

\begin{itemize}
\item For the aggressive driver:

\begin{equation}
\lambda_{agg} = \frac{K_{cor,agg}}{\sum K_{all,agg}}, \quad 0\leqslant\lambda_{agg} \leqslant 1
\end{equation}

\item For the moderate driver

\begin{equation}
\lambda_{mod} = \frac{K_{cor,mod}}{\sum K_{all,mod}}, \quad 0 \leqslant \lambda_{mod} \leqslant 1
\end{equation}

\end{itemize}
where $ K_{\star,\bullet} $ denotes the \{$ \star $\} number of \{$ \bullet $\} driving pattern. $ \star \in \{ cor,all \} $, $ \bullet \in \{ agg,mod \} $ and the symbol $ cor $ represents the correct choice. For example, $ K_{cor,mod} $ represents the number of the clustering points that are correctly classified into the moderate pattern region for the moderate driver. 

\subsection{Testing Results and Analysis}

For different testing data sets $ \tilde{\boldsymbol{\eta}} $, the test results with corresponding to training data sets $ \boldsymbol{\eta} $ are conducted using off-line and on-line method. Then, the comparisons between $ k $MC-SVM and SVM are presented. 

\subsubsection{Off-line Testing}
\begin{figure}[thpb]
\centering
\includegraphics[scale=0.65]{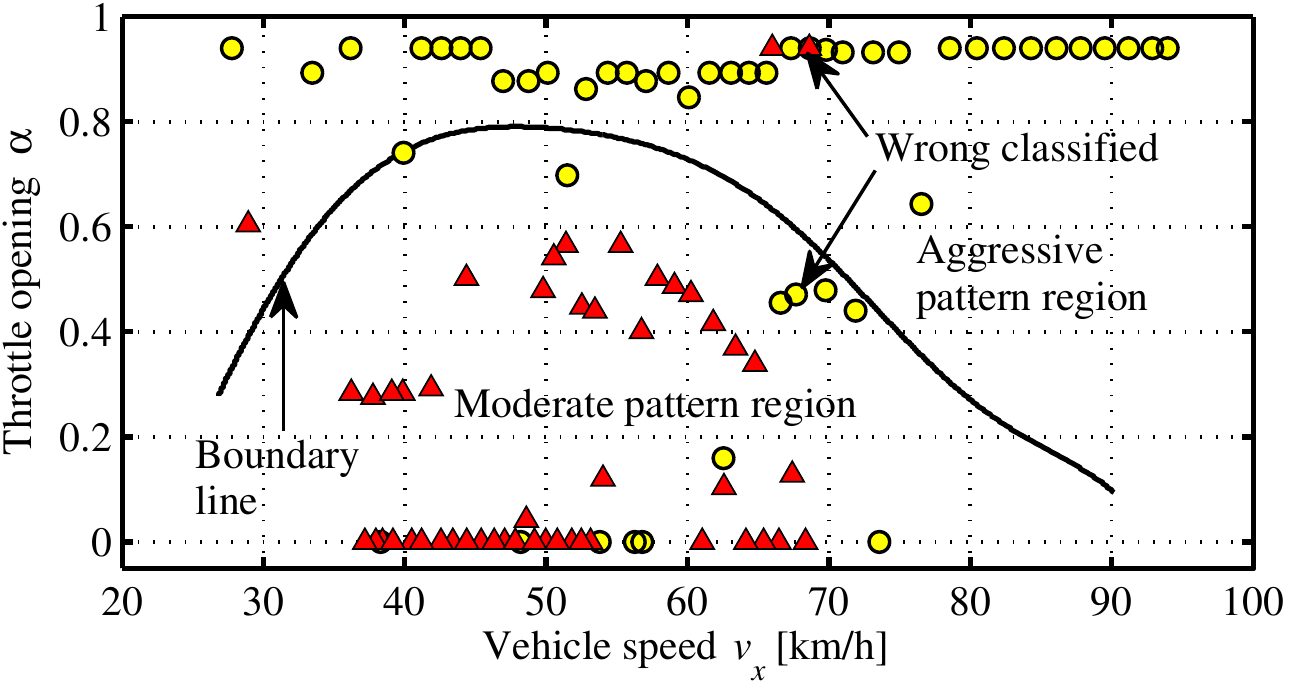}
\caption{The testing evaluation for two types driver using the $ K $MC-SVM method based on training data in Fig. \ref{figure5}.}
\label{figure9-1}
\end{figure}

\begin{figure}[thpb]
\centering
\includegraphics[scale=0.6]{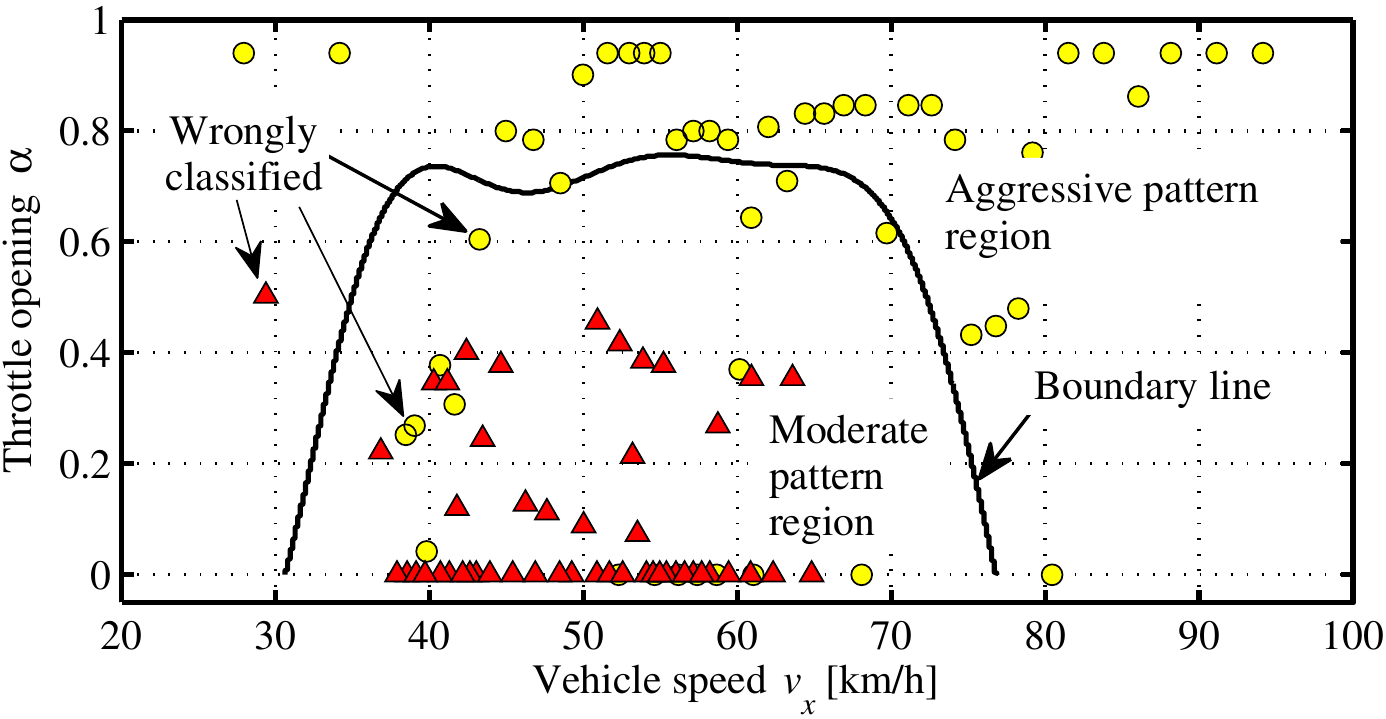}
\caption{Classification results for two types driver using $ k $MC-SVM method.}
\label{figure10-1}
\end{figure}

For off-line test, all the data are recorded and clustered first, and then the clustered testing data is used for evaluation. The off-line test results of testing data corresponding to the training data sets in Fig. \ref{figure5} and Fig. \ref{figure5-1} are shown in Fig. \ref{figure9-1}, Fig.\ref{figure10-1}, and Table \ref{table_1}. The symbol $ \lambda_{\bullet,1} $ denotes the first group of drivers with $ \{ \bullet \} $ driving  pattern.
From the Table \ref{table_1}, it is obvious that the $ k $MC-SVM method has more benefits than the traditional SVM method on the recognition of driving styles.  For the first group of drivers, the $ k $MC-SVM method is more efficient to recognizing the aggressive driver than SVM method, but not improvement on recognizing the moderate drivers.  For the second group of drivers, the $ k $MC-SVM method has greatly benefits for recognizing both driving patterns, improve recognition accuracy by $ 5.49\% $ and $ 11.20\% $.

\subsubsection{On-line Testing}
\begin{figure}[thpb]
\centering
\includegraphics[scale=0.6]{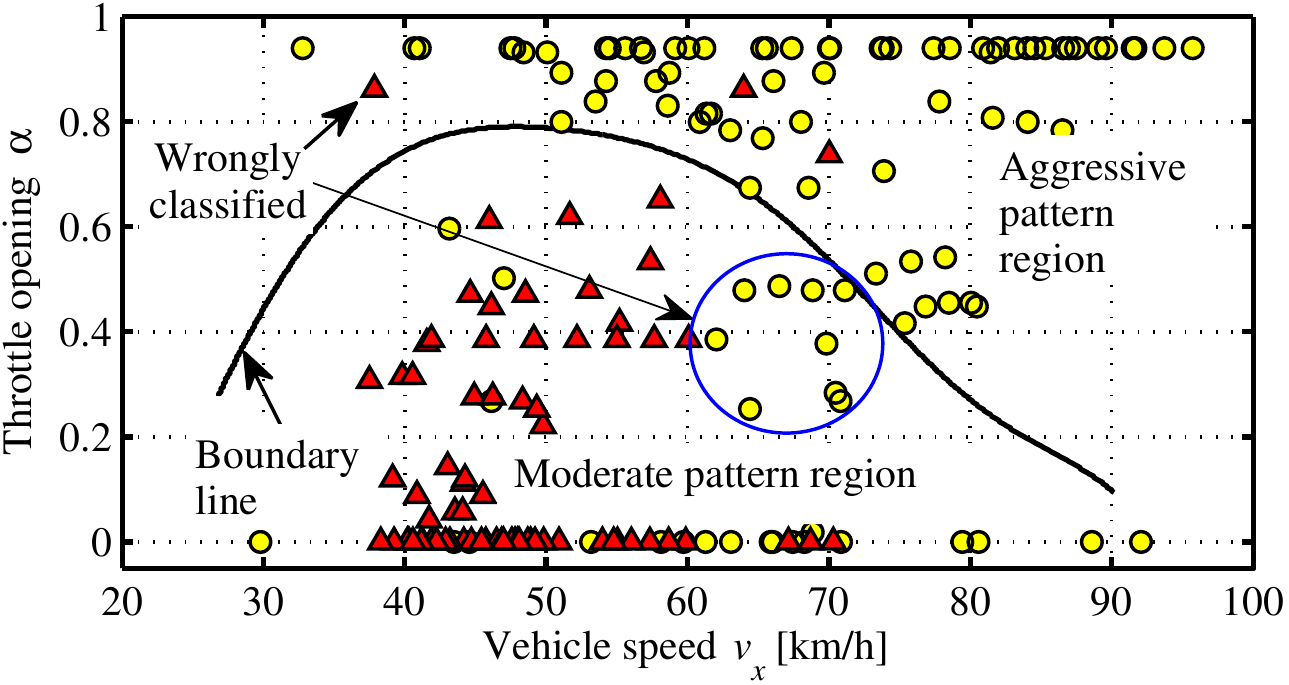}
\caption{The on-line testing evaluation for two types driver using the $ k $MC-SVM method. 1) yellow dots: aggressive driving testing-data; 2) red triangle: moderate driving testing-data.}
\label{figure9-2}
\end{figure}

\begin{figure}[thpb]
\centering
\includegraphics[scale=0.55]{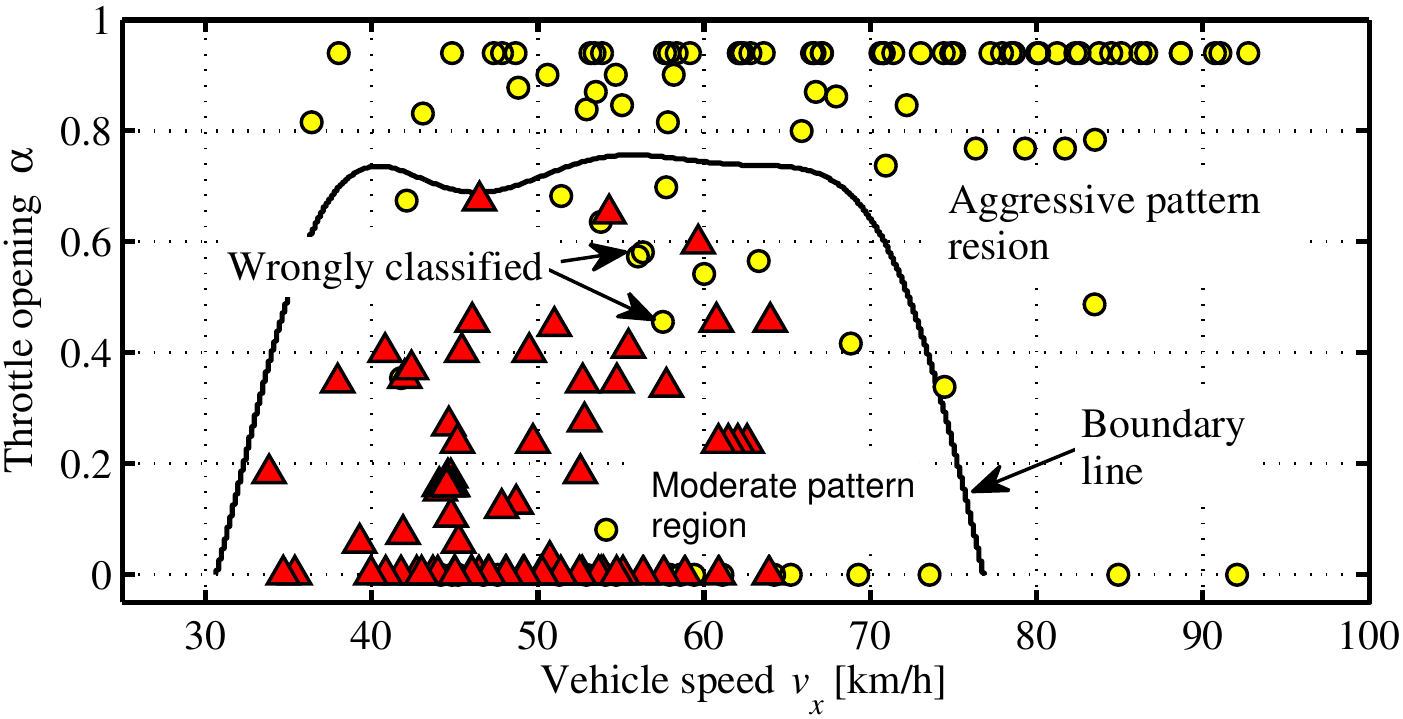}
\caption{The on-line testing evaluation for two types driver using the cross-validation method. 1) yellow dots: aggressive driving testing-data; 2) red triangles: moderate driving testing-data.}
\label{figure10-2}
\end{figure}

We assume that the human driver's driving pattern are treated as constant in a fixed time interval $ [t,t+\tau] $, $ t $ is the current time and $ \tau $ is the fixed time span. In this paper, the past information or driving pattern of the human driver during a fixed time span $ \tau $ is adopted to represent the current driving pattern. Therefore, the current driving style is described as follows:

\begin{equation}
P_{t}=\hbar(\boldsymbol{\eta}_{t-\tau,t},y_{t-\tau,t})
\end{equation}
where $ \boldsymbol{\eta}_{t-\tau,t} $ is the clustering point based on the past time span $ \tau $ and $ \tau=1.4s $. Therefore, the data sequence of driving data can be clustered in the time span $ [0,\tau], [\tau,2\tau],\dots,[(K-1)\tau,K\tau] $, generating $ K $ clustering data sets. The testing results are shown in Fig. \ref{figure9-2}, Fig. \ref{figure10-2}, and Table \ref{table_1}.
According to the Table \ref{table_1}, we can make conclusion that the $ k $MC-SVM method has a higher accuracy for recognition of driving patterns than SVM. For the first group of drivers, the accuracy of recognition is improved by $ 17.23\% $ and $ 1.76\% $ for the aggressive driver and moderate driver, respectively. For the second group, the accuracy of recognition is improved by $  9.78\% $ and $ 17.06\% $ for the aggressive and moderate driver, respectively.

\begin{table}[ht]
\caption{Results for Testing Data with the Number of Clusters $ K=\sqrt{n/3} $, Time Span $ \tau=1.4s $}
\label{table_1}
\begin{center}
\begin{tabular}{c|c|c|c|c|c}
\hline
\hline
  &  Method    & $ \lambda_{agg,1} $  & $ \lambda_{mod,1} $ & $ \lambda_{agg,2} $ & $ \lambda_{mod,2} $ \\
\hline
\multirow{2}{*}{$ k $MC-SVM} & off-line &  77.78\%  &  92\%  &  79.49\%  & 95\%  \\
& on-line &  87.04\%  &  94.92\%  &  82.72\%  &   100\% \\
\hline
SVM & off-line & 74.25\% & 93.28\% & 75.35\% & 85.43\%\\
\hline
\hline
\end{tabular}
\end{center}
\end{table}

\subsection{Time-Cost Analysis}
In this section, to show the time-saving benefits of $ k $MC-SVM, one data set for testing is selected from the all of the datasets, and  then the remaining data sets are used for training. The training time  is listed in Table. \ref{table_2} using $ k $MC-SVM and SVM, where $ n \in \{ n_{training},n_{testing} \}$, and $ n_{training}= 112,383 $, $ n_{testing}=6,177 $, $ T $[s] is the calculation time. For $ k $MC-SVM method, $ T $[s] consists of the clustering time and training time, and for SVM method, $ T $[s] is the training time. Because the testing time is very short, we neglected it here. According to Table \ref{table_2}, we can know that the $ k $MC-SVM method can shorten time greatly, though the on-line recognition efficiency is down slightly, compared with the SVM method. 

\begin{table}[ht]
\caption{Comparison of the Time-cost of Recognition by Adopting $ k $MC-SVM and SVM}
\label{table_2}
\begin{center}
\begin{tabular}{c|c|c|c||c|c|c}
\hline
\hline
\multirow{2}{*}{ $ K $ }    & \multicolumn{3}{|c||}{$ k $MC-SVM}  &  \multicolumn{3}{|c}{SVM} \\
\cline{2-7} & $T$[s] & $ \lambda_{agg} $ & $ \lambda_{mod} $ & $T$[s] & $ \lambda_{agg} $ & $ \lambda_{mod} $\\
\hline 
$ \sqrt{n/2} $ & 169.35 & 78.89\% & 82.07\% & 740.18 & 85.46\% & 88.19\%\\
\hline
\hline
\end{tabular}
\end{center}
\end{table}

\section{Conclusion}

A rapid pattern-recognition method, called $ k $MC-SVM, is developed by combining the $ k $-means clustering and SVM, and subsequently applied to recognize driver's curve-negotiating patterns, i.e., aggressive and moderate. The kMC-SVM, compared with SVM, can not only shorten the recognition time but improve the recognition for classification issues that are not linearly separable. First, to reduce the number of support vectors, the $ k $-means clustering method is applied, clustering the original data sets into $ K $ subsets. And then, based on the clustering results, the SVM is applied to generate the hyperplane for datasets with different labels. Last, the cross-validation experiments is designed to shown the benefits of the proposed method. The testing results show that the $ k $MC-SVM is able to not only shorten the training time for classification model, but improve recognition, compared with SVM method. 





\end{document}